\documentclass{article}

\usepackage{microtype}
\usepackage{graphicx}
\usepackage{subfigure}
\usepackage{booktabs} 
\usepackage{gensymb}
\usepackage{amssymb}
\usepackage{caption}

\usepackage{hyperref}

\usepackage[accepted]{icml2021}

\icmltitlerunning{Estimation of Air Pollution from Remote Sensing Data}

\begin{document}

\twocolumn[

\icmltitle{Estimation of Air Pollution with Remote Sensing Data: \\ 
           Revealing Greenhouse Gas Emissions from Space}



\icmlsetsymbol{equal}{*}

\begin{icmlauthorlist}
\icmlauthor{Linus Scheibenreif}{stg}
\icmlauthor{Michael Mommert}{stg}
\icmlauthor{Damian Borth}{stg}
\end{icmlauthorlist}

\icmlaffiliation{stg}{Institute of Computer Science, University of St. Gallen, Switzerland}
\icmlcorrespondingauthor{Linus Scheibenreif}{linus.scheibenreif@unisg.ch}

\icmlkeywords{Remote Sensing, Air Pollution, Deep Learning, NO2, Machine Learning, ICML}

\vskip 0.3in
]


\printAffiliationsAndNotice{}

\begin{abstract}
Air pollution is a major driver of climate change. Anthropogenic emissions from the burning of fossil fuels for transportation and power generation emit large amounts of problematic air pollutants, including Greenhouse Gases (GHGs).
Despite the importance of limiting GHG emissions to mitigate climate change, detailed information about the spatial and temporal distribution of GHG and other air pollutants is difficult to obtain. Existing models for surface-level air pollution rely on extensive land-use datasets which are often locally restricted and temporally static. This work proposes a deep learning approach for the prediction of ambient air pollution that only relies on remote sensing data that is globally available and frequently updated. Combining optical satellite imagery with satellite-based atmospheric column density air pollution measurements enables the scaling of air pollution estimates (in this case NO$_2$) to high spatial resolution (up to $\sim$10m) at arbitrary locations and adds a temporal component to these estimates. The proposed model performs with high accuracy when evaluated against air quality measurements from ground stations (mean absolute error $<$6$~\mu g/m^3$). Our results enable the identification and temporal monitoring of major sources of air pollution and GHGs.
\end{abstract}

\begin{figure}[t]
\vskip 0.2in
\begin{center}
\centerline{\includegraphics[width=\columnwidth]{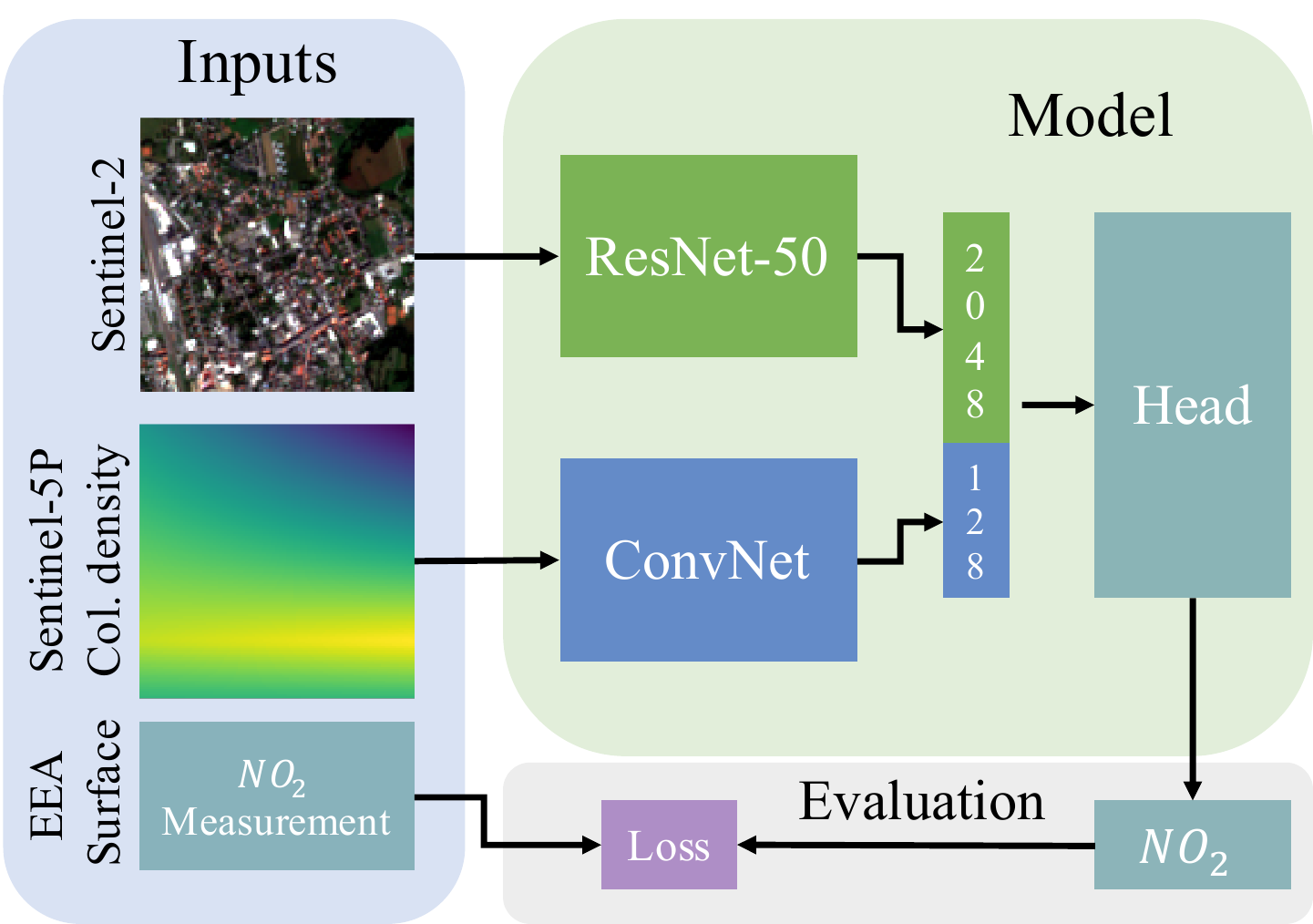}}
\caption{Overview of the proposed air pollution prediction system.}
\label{fig:model}
\end{center}
\vskip -0.2in
\end{figure}

\section{Introduction}
\label{section:introduction}

\begin{figure*}[t]
    \centering
\includegraphics[width=\textwidth]{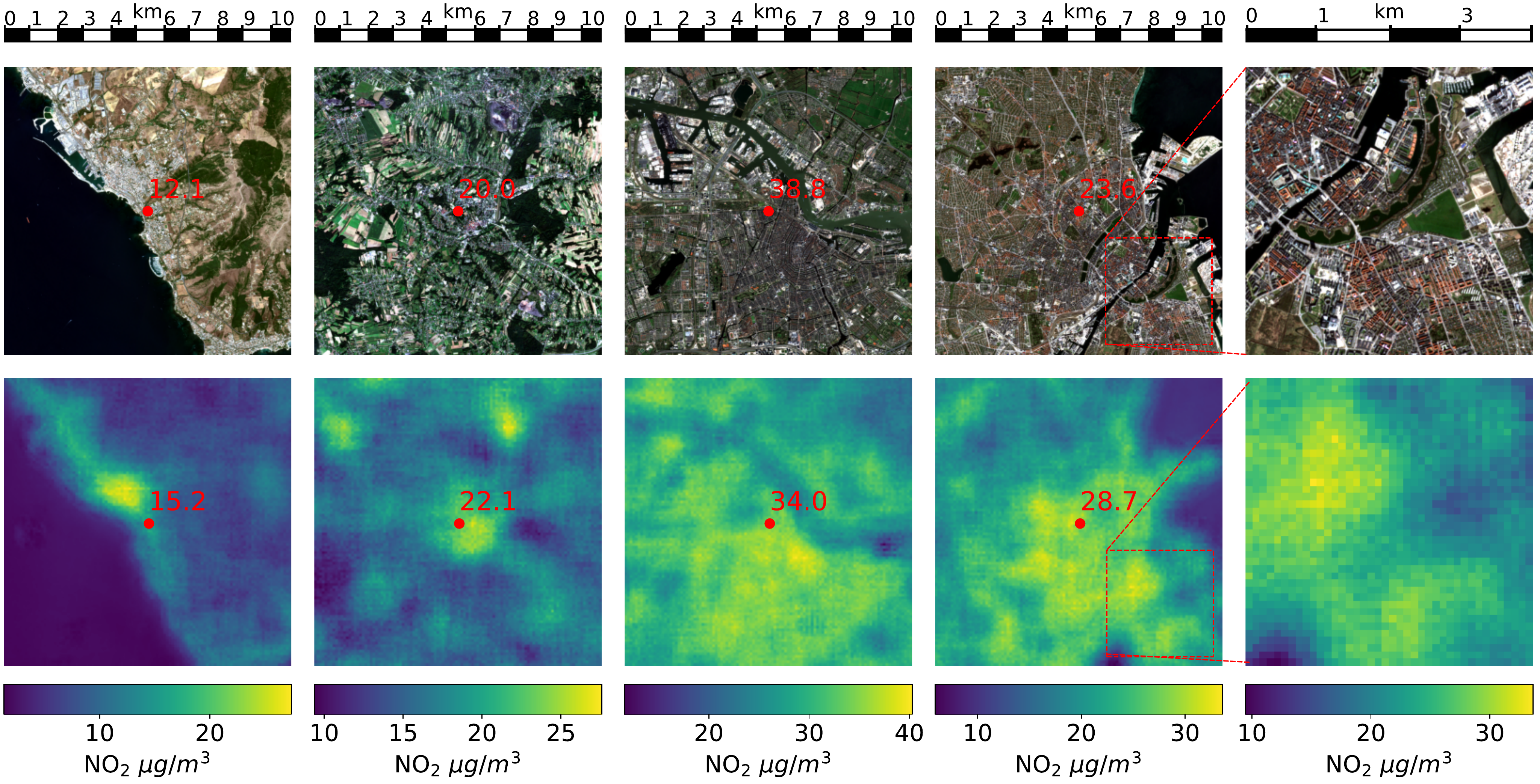}
\vskip -0.13in
    \caption{Exemplary NO$_2$ predictions based on Sentinel-2 and Sentinel-5P input data. \textbf{Top:} RGB bands of the Sentinel-2 image, red dots mark locations of air quality stations, red text indicates the average NO$_2$ concentration measured on the ground during the 2018-2020 timespan. \textbf{Bottom:} Predicted NO$_2$ concentrations for the locations above (not seen during training) with predictions at the exact position of air quality stations in red. The heatmaps are constructed from individual predictions for overlapping 120$\times$120 pixel tiles of the top image and corresponding Sentinel-5P data, resulting in an effective spatial resolution of 100m. This approach is equally applicable to locations without air quality stations, providing a means to map air pollution on the surface level to identify sources of air pollution and GHG emissions (see Fig. \ref{fig:A1} for more examples).}
    \label{fig:maps}
\end{figure*}


Air pollution and the emission of GHGs is the main cause of climate change with annual global emission levels still on the rise \cite{friedlingstein2019global}. In particular, anthropogenic GHG emissions from the combustion of fossil fuels in industrial plants or for transportation are harmful to the environment and contribute to global warming trends \cite{ledley1999climate}. Besides the primary greenhouse gas, CO$_2$, the burning of fossil fuels also emits molecules like NO$_2$ and CO, which have been used as proxy for the estimation of CO$_2$ emissions \cite{berezin2013multiannual}.
Detailed information about sources and distribution of air pollutants within the atmosphere is of high relevance for a number of applications with climate change impact, including the compilation of emission inventories \cite{eggleston2006}, the design and implementation of pollution limits \cite{bollen2014air}, and the quantification of large anthropogenic emissions \cite{liu2020methodology}.

At present, continual data on air pollution concentrations in the atmosphere are primarily collected through two different approaches with distinct drawbacks. On the Earth's surface, networks of measurement stations record the concentration of various chemicals at select locations \cite{guerreiro2014air}. Such networks are commonly run by environmental agencies and provide frequent measurements while often lacking in spatial coverage. This drawback can be partly addressed by space-borne air pollution monitoring: satellites equipped with spectrometers measure the abundance of select molecules in the form of atmospheric column densities \cite{gupta2006satellite}. While their position in Earth's orbit allows satellites to frequently map most locations on Earth, remote sensing spectrometers currently only provide spatial resolutions in the kilometer range and with little information about the pollutant's vertical distribution. Specifically, the estimation of concentrations near the surface, where these pollutants originate from, is a non-trivial task \cite{scheibenreif2021NO2}.
One of the primary anthropogenic air pollutants is Nitrogen Dioxide (NO$_2$). Elevated levels of NO$_2$ harm the vegetation, contribute to acid rain, and act as a precursor of potent GHGs like Ozone \cite{montzka2011non}. Additionally, NO$_2$ is jointly emitted with CO$_2$ during the combustion of fossil fuels at high temperatures, making it a suitable proxy to identify CO$_2$ emission sources \cite{konovalov2016estimation, goldberg2019exploiting}. This work leverages a large body of publicly available NO$_2$ concentration measurements on the ground by the European Environment Agency's\footnote{\url{eea.europa.eu}} (EEA) network of air quality stations and satellite measurements from the European Space Agency's (ESA) Copernicus program to investigate the distribution of air pollutants through a deep learning approach. The results of this work enable the identification of major sources of GHG emissions and their temporal monitoring on a global scale.

\section{Background}

\begin{table*}[t]
\caption{Performance metrics for NO$_2$ estimation with various model architectures and datasets on unseen observations, averaged over 10 training runs with varying random seeds (PT: Pre-trained model, *-T10: Performance of top model out of 10 training runs).}
\label{table:results}
\vskip 0.15in
\begin{center}
\begin{small}
\begin{sc}
\begin{tabular}{llcccccccc}
\toprule
 Data & Time & N-Obs. & PT & R2 & R2-T10 & MAE & MAE-T10 & MSE & MSE-T10 \\
\midrule
Sen.-2         & 2018-20    & 3.2k    & $\times$ & 0.25$\pm$0.05 & 0.28 & 8.06$\pm$0.49 & 7.31 & 105.7$\pm$10.29 & 91.72\\
Sen.-2         & 2018-20    & 3.2k    & \checkmark  & 0.45$\pm$0.03 & 0.49 & 6.62$\pm$0.17 & 6.23 & 77.03$\pm$3.64 & 65.81 \\
Sen.-2,5P      & 2018-20    & 3.1k    & $\times$ & 0.38$\pm$0.03 & 0.43 & 7.06$\pm$0.35 & 6.68 & 83.72$\pm$4.14 & 78.4\\
Sen.-2,5P      & 2018-20    & 3.1k   & \checkmark  & \textbf{0.54}$\pm$\textbf{0.04} & \textbf{0.59} & \textbf{5.92}$\pm$\textbf{0.44} & \textbf{5.42} & \textbf{62.52}$\pm$\textbf{5.47} & \textbf{56.28}\\
Sen.-2,5P      & Quart.   & 19.6k   & \checkmark  & 0.52$\pm$0.05 & 0.57 & 6.24$\pm$0.22 & 5.98 & 73.1$\pm$6.88 & 66.12 \\
Sen.-2,5P      & Month.    & 59.6k   & \checkmark & 0.51$\pm$0.01 & 0.53 & 6.54$\pm$0.15 & 6.31 & 78.96$\pm$4.2 & 73.74 \\
\bottomrule
\end{tabular}
\end{sc}
\end{small}
\end{center}
\vskip -0.1in
\end{table*}

Various prediction and interpolation techniques have been used to derive detailed information about the spatial distribution of air-borne pollutants such as GHGs. Typically, these models are based on point measurements from air quality monitoring stations that are spatially limited to specific locations. 
Beyond interpolation with geostatistical approaches like kriging \cite{janssen2008spatial}, land-use-regression (LUR) is commonly applied to incorporate covariates such as population density or traffic data into the models \citep[see][for a review]{hoek2008review}. LUR models often involve variable selection procedures to identify predictive inputs over large sets of candidate variables, making it difficult to scale to regions not covered by detailed datasets, even if some air quality measurements are available. Building on existing work that incorporates satellite measurements into LUR frameworks \cite{novotny2011national}, we extend this approach to model air pollution at high spatial resolution solely from satellite data.
Our work is based on NO$_2$ concentration measurements by the EEA. We consider NO$_2$ as pollutant of interest due to its relevance as major anthropogenic air pollutant and chemical properties that facilitate its detection from space with high accuracy (opposed to GHGs like CO$_2$). Additionally, it is co-emitted with CO$_2$ in the burning of fossil fuels, which makes it possible to constrain CO$_2$ emissions from NO$_2$ measurements \cite{berezin2013multiannual}.
To facilitate the identification of air pollutant sources, which are commonly located on the ground, we model surface-level concentrations (rather than e.g. atmospheric column densities). The EEA network of air quality stations provides frequent (mostly hourly) measurements of NO$_2$ concentrations at more than 3,000 locations in Europe.
Additionally, remote sensing data from ESA's Sentinel-2 and Sentinel-5P satellites is utilized to model air quality. Sentinel-2 is a constellation of two satellites carrying the Multi Spectral Instrument, a spectrometer covering the visible, near-infrared and shortwave-infrared wavelengths with imaging resolutions between 10 and 60 meters \cite{drusch2012sentinel}. Sentinel-2 data is widely used in applications like land cover classification or crop monitoring \cite{helber2019eurosat} but also for the monitoring of GHGs at locations of interest \citep[e.g., based on the presence of smoke plumes,][]{mommert2020characterization}. In our work, globally available and continually updated Sentinel-2 images replace conventional LUR predictor variables such as street networks, population density or vegetation information.
The Sentinel-5P satellite observes trace-gases and aerosols in the atmosphere through differential optical absorption spectroscopy \cite{veefkind2012tropomi}. It provides daily global coverage for gases including NO$_2$, O$_3$, CO or CH$_4$ with a spatial resolution of $5\times3.5$~km. We utilize the NO$_2$ tropospheric column density product of Sentinel-5P to model the temporal variation in surface NO$_2$ levels.

\section{Methods}

This work approaches the estimation of air pollution as a supervised computer vision problem. We collect a dataset of harmonized remote sensing data from Sentinel-2 and Sentinel-5P, spatially and temporally aligned with measurements from air quality monitoring stations. The proposed model is trained on pairs of remote sensing input and air quality target values (see Fig. \ref{fig:model}), which yields a system that predicts air pollution levels solely from globally available remote sensing data\footnote{code available at \url{github.com/HSG-AIML/RemoteSensingNO2Estimation}}.

\subsection{Data Processing}
We consider the 2018-2020 timespan, historically limited by the start of the Sentinel-5P nominal mission.
NO$_2$ measurements by EEA air quality stations are filtered to remove values with insufficient quality (\texttt{validity} or \texttt{verification} value $\neq$1). Besides modelling the entire 2018-2020 timespan, we also investigate the possibility to estimate NO$_2$ concentrations at quarterly and monthly frequencies. To that end, the mean of NO$_2$ measurements for each frequency is used as prediction target. To build the dataset, we downloaded Sentinel-2 Level-2A data (i.e. corrected for atmospheric effects and enriched with cloud masks) with low cloud-coverage at the locations of air quality stations, containing 12 different bands (band 10 is empty in the case of Level-2A data). The images were then cropped to 120$\times$120 pixel size ($\sim$1.2$\times$1.2~km) centered at the location of interest, and all bands were upsampled to 10~m resolution with bilinear upsampling. Additionally, we visually inspected the RGB bands of all images to ensure that no clouds or artifacts are present.
Similarly, Sentinel-5P data over Europe was downloaded for the 2018-2020 timespan (5449 Level-2 products) and mapped to a common rectangular grid of 0.05$\times$0.05~$\degree$ ($\sim$5$\times$5~km) resolution after removing invalid measurements (\texttt{qa\_value} $<$75). The resulting dataset was averaged at the different temporal frequencies and 20$\times$20~km regions at the locations of air quality stations were extracted. To facilitate processing despite the coarse resolution ($\sim$500$\times$ lower than Sentinel-2), we linearly interpolated the Sentinel-5P data to 10~m resolution and cropped to 120$\times$120 pixel centered at the locations of interest.

\subsection{Model Architecture}
Our core model for NO$_2$ prediction from imaging data is based on the ResNet-50 architecture \cite{he2016deep} (see Fig. \ref{fig:model}). The input layer is modified to accommodate the 12-band Sentinel-2 input data and the final layer is replaced by two dense layers with ReLU activation (named \texttt{head}) that map the 2048-dimensional feature vector to a scalar value. We employ transfer learning by pretraining the model on a land-cover classification (LCC) task with the BigEarthNet dataset \cite{sumbul2019bigearthnet}. After pretraining, the final classification layer is replaced by the \texttt{head}, i.e., only the trained convolutional backbone of the ResNet is retained. Intuitively, learned features that are informative for LCC (e.g., distinguishing industrial areas from forests) will also be useful when estimating emission profiles of different areas.
To handle additional input data from Sentinel-5P, the model architecture is extended with a small sub-network, consisting of two convolutional layers (with \texttt{10,15} channels and kernel sizes \texttt{3,5}, respectively), each followed by ReLU activation functions and max-pooling (kernel size \texttt{3}), and a final linear layer. This sub-network is much smaller than the ResNet-50 used to process the Sentinel-2 input stream to reflect the lower native resolution and single band nature of the Sentinel-5P data. It learns a 128 dimensional latent vector from the Sentinel-5P input image. To obtain an NO$_2$ prediction, the latent vectors of both input-streams are concatenated and again processed by the \texttt{head} with adjusted input dimensions (2048+128).
All presented models were trained 10 times with varying seeds, mean-squared-error loss function and random train/test/validation split of 60:20:20. To limit overfitting, training is stopped once the loss on the validation set stops decreasing. Additionally, we employ random flipping and rotation of the inputs as augmentation during training.

\section{Experiments}
\begin{figure}
    \centering
    \includegraphics[width=\columnwidth]{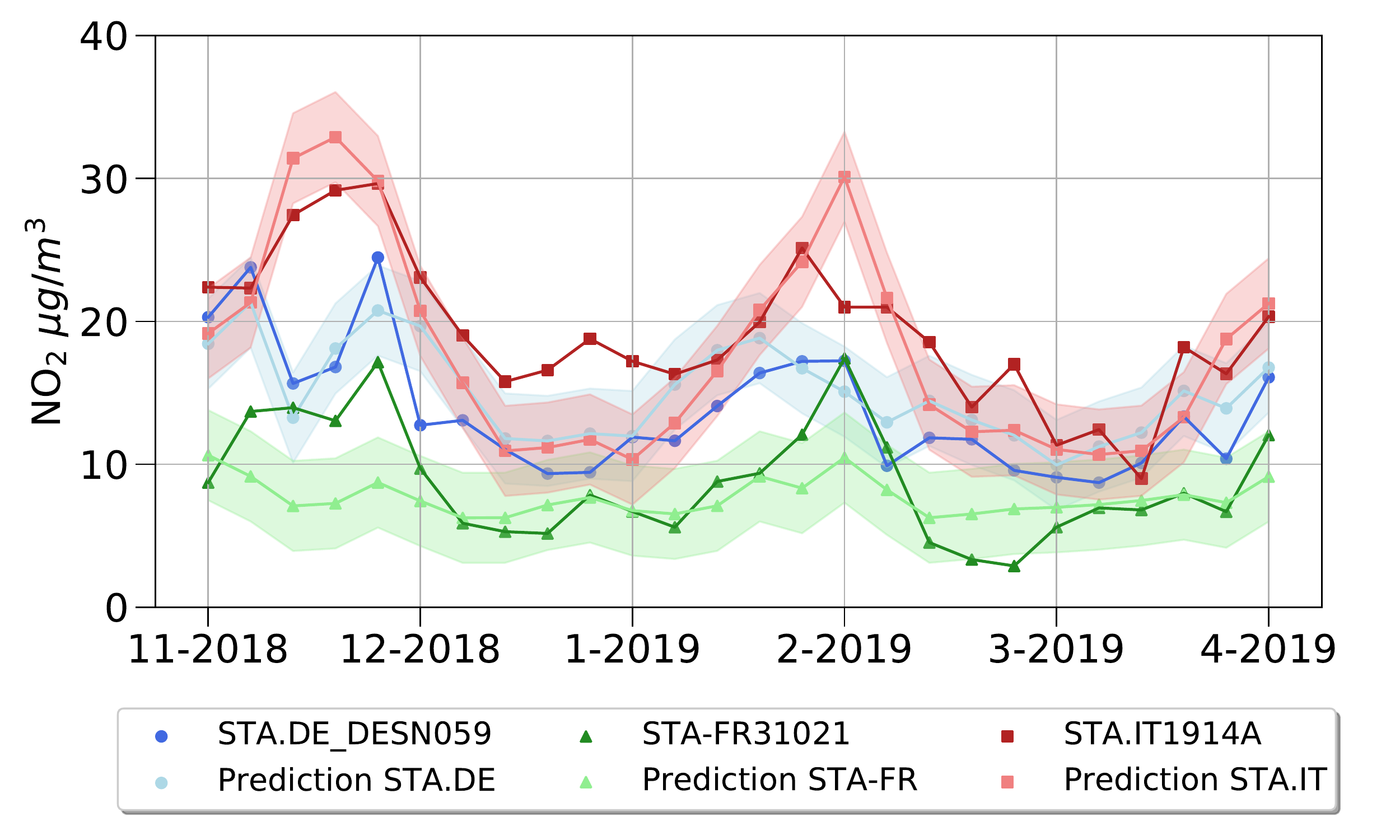}
    \vskip -0.15in
    \caption{Monthly average NO$_2$ measurements from three EEA air quality stations in Germany, France and Italy (dark colors) and monthly NO$_2$ predictions based on Sentinel-2 and Sentinel-5P measurements at the same locations (not seen during training). The shaded area indicates the model's MAE envelope centered at the nominal predictions.}
    \label{fig:temporal}
\vskip -0.2in
\end{figure}

To assess the predictive power of Sentinel-2 images for air pollution prediction we initially train a model on Sentinel-2 images as inputs with air quality station measurements as target. Using only Sentinel-2 images forces the model to associate features that are apparent in medium-resolution satellite imagery, like built-up areas, forests or streets, with representative NO$_2$ levels.
Training this model from scratch leads to a mean-absolute-error (MAE) of 8.06$\pm$0.49~$\mu g/m^3$ and R2-Score of 0.25$\pm$0.05~$\mu g/m^3$ (see Table \ref{table:results}), presumably limited by the dataset size of only 3,227 images. 
Following the intuition that LCC shares predictive features with air pollution prediction, we then investigated a transfer learning approach by pre-training the ResNet backend on BigEarthNet \citep[590,326 images with multi-label annotations,][]{sumbul2019bigearthnet}. Using the pretrained backend in the NO$_2$ prediction model and fine-tuning on the Sentinel-2 images at air quality stations, we obtain a significantly better performance. The MAE drops to 6.62$\pm$0.17~$\mu g/m^3$ with an R2-Score of 0.45$\pm$0.03. This first result supports our hypothesis that medium-resolution satellite imagery is valuable for the estimation of ambient air pollution.
We then investigated ways of incorporating tropospheric column density measurements of NO$_2$ from Sentinel-5P into the model using a second input stream. The additional satellite data results in a further performance increase with an MAE of 5.92$\pm$0.44 and R2-Score of 0.54$\pm$0.04 and allows us to derive detailed pollution maps for any location of interest (see Fig. \ref{fig:maps}). Inclusion of Sentinel-5P data, which is updated daily, also provides us with a way of modeling temporal variations in NO$_2$ levels. Aggregating the data at higher frequency significantly increases the number of observations (from 3.1k to 19.6k quarterly and 59.6k monthly samples), which enables the model to maintain a performance comparable to the static predictions (MAE of 6.24$\pm$0.22 and 6.52$\pm$0.15~$\mu g/m^3$ for quarterly and monthly predictions, respectively). Similarly, the R2-Scores remain at 0.52$\pm$0.05 (quarterly) and 0.51$\pm$0.01 (monthly) despite the increase in prediction frequency. This makes it possible to model seasonal changes in NO$_2$ concentrations with good accuracy (see Fig. \ref{fig:temporal}).

\section{Conclusion}
We present an end-to-end approach for the estimation of surface NO$_2$ concentrations with deep learning. Utilizing only remote sensing data as inputs, it is possible to model arbitrary regions on Earth, independent of the availability of detailed datasets as commonly used in the prediction of air pollutant distributions. Qualitative evaluation shows that our models are robust across most regions of Europe, except for rare atypical locations that are badly represented in our dataset, e.g., snowy mountain peaks. In future work, measurements from air quality networks outside of Europe can be incorporated into model training to improve model generalization. 

The focus of this work on NO$_2$ allows us to leverage a large corpus of pollutant measurements from air quality stations and from space to better localize the sources of air pollution and GHG emitters. This information enables an approximate analysis of the spatial and temporal distribution of air pollutants and GHG emissions alike, providing constraints that are vital for our effort to reduce GHG emissions and reaching the net-zero emission target.

\section*{Acknowledgements}
We thank the ESA Copernicus programme and the European Environment Agency for providing the data used in this work.



\bibliography{references}
\bibliographystyle{icml2021}

\newpage\null\thispagestyle{empty}\newpage
\appendix
\section{Appendix}
\renewcommand{\thefigure}{A\arabic{figure}}
\setcounter{figure}{0}
\centering
\begin{minipage}{\textwidth}
\makebox[\textwidth]{
  \includegraphics[width=\textwidth]{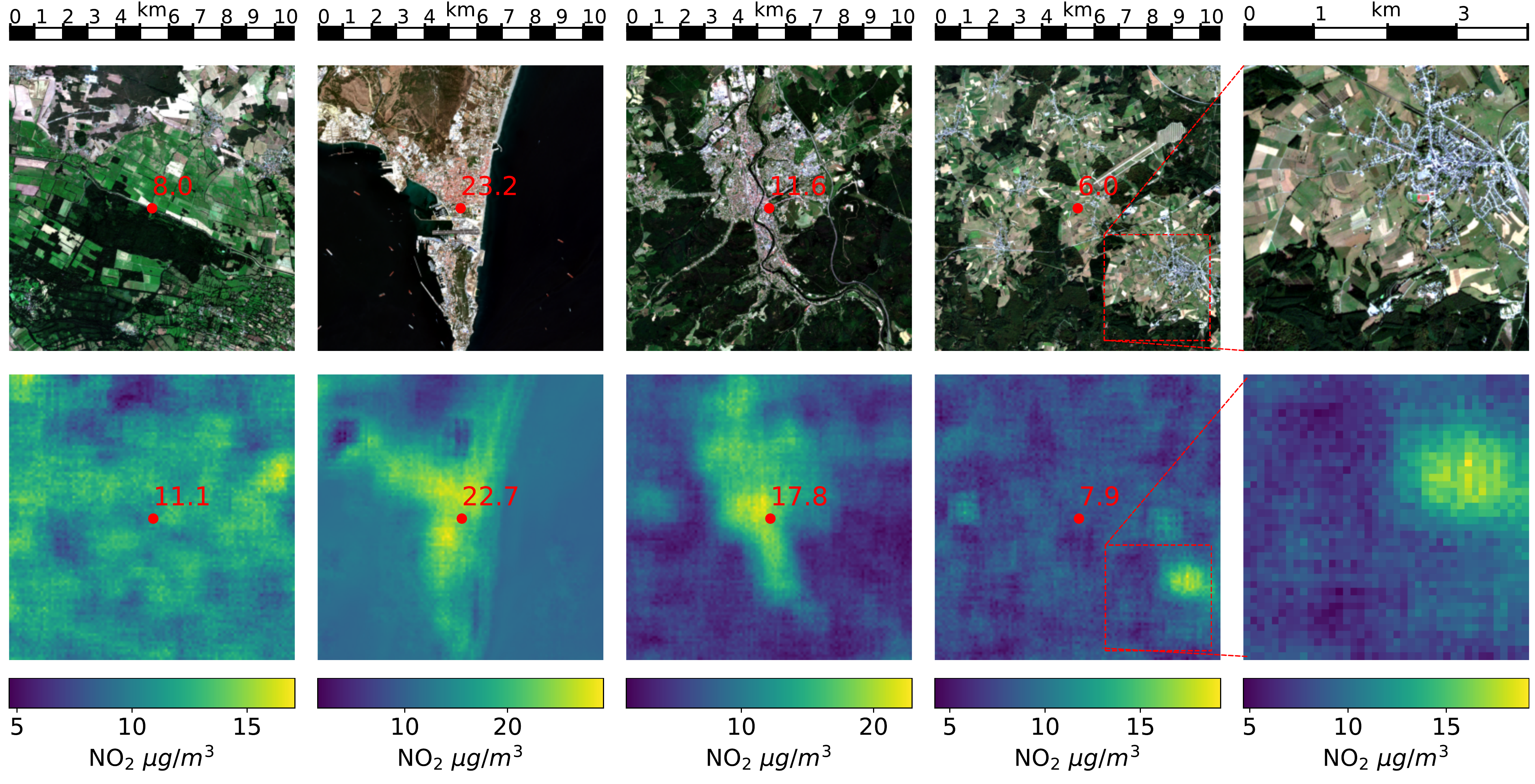}}
  \vskip 0.2in
 \makebox[\textwidth]{%
  \includegraphics[width=\textwidth]{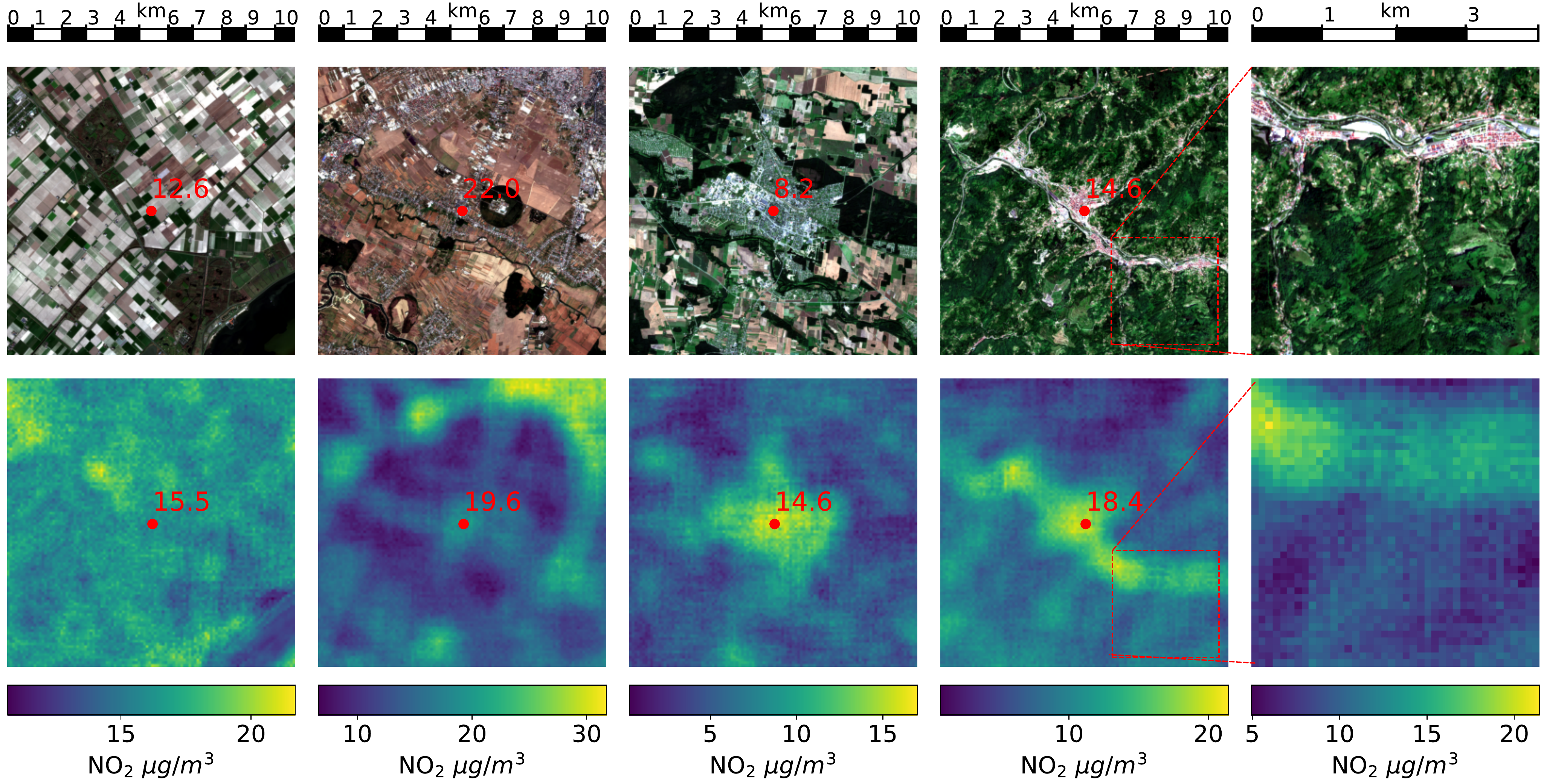}
  }
\captionof{figure}{Additional examples of surface NO$_2$ predictions from Sentinel-2 and Sentinel-5P data across Europe. Pictures are centered at locations of EEA air quality stations (red dots). The red text indicates average NO$_2$ measurements (RGB images) and corresponding NO$_2$ estimates (heatmaps).}
\label{fig:A1}
\end{minipage}

\end{document}